%% file: main.tex
\documentclass{article}
\usepackage{ijcai22}

\usepackage{times}

\usepackage{soul}
\usepackage{url}
\usepackage[hidelinks]{hyperref}
\usepackage[utf8]{inputenc}
\usepackage[small]{caption}
\usepackage{graphicx}
\usepackage{amsmath}
\usepackage{booktabs}
\usepackage{color}
\usepackage{tabularx}
\usepackage{subcaption}
\usepackage[table]{colortbl}
\usepackage{xcolor}

\title{Crop Type Identification for Smallholding Farms:\\ Analyzing Spatial, Temporal and Spectral Resolutions in Satellite Imagery}

\author{
Depanshu Sani$^1$\footnote{Equal contribution}\and
Sandeep Mahato$^{2*}$\and
Parichya Sirohi$^1$\and
Saket Anand$^1$\and \\
Gaurav Arora$^{1}$\and
Charu Chandra Devshali$^{2}$\And
T. Jayaraman$^2$\\
\affiliations
$^1$Indraprastha Institute of Information Technology, Delhi, India\\
$^2$M S Swaminathan Research Foundation, Chennai, India\\
\emails
depanshus@iiitd.ac.in
}
    
\begin{document}

\maketitle

\input{sections/abstract}

\input{sections/introduction}

\input{sections/data_collection}

\input{sections/methodology}

\input{sections/experiments}

\input{sections/results}

\input{sections/conclusion}

\clearpage
\bibliographystyle{named}
\bibliography{ijcai22}

\end{document}

%% file: sections/abstract.tex
\begin{abstract}
The integration of the modern Machine Learning (ML) models into remote sensing and agriculture has expanded the scope of the application of satellite images in the agriculture domain. In this paper, we present how the accuracy of crop type identification improves as we move from medium-spatiotemporal-resolution (MSTR) to high-spatiotemporal-resolution (HSTR) satellite images. We further demonstrate that high spectral resolution in satellite imagery can improve prediction performance for low spatial and temporal resolutions (LSTR) images. The F1-score is increased by $7\%$ when using multispectral data of MSTR images as compared to the best results obtained from HSTR images. Similarly, when crop season based time series of multispectral data is used we observe an increase of $1.2\%$ in the F1-score. The outcome motivates further advancements in the field of synthetic band generation.
\end{abstract}

%% file: sections/introduction.tex
\section{Introduction}
This paper demonstrates the challenges for crop type identification in the Cauvery Delta in Tamil Nadu, India and potential approaches to address the challanges. Paddy is the principal crop in the Delta, cultivated once or twice annually, responsible for the food security and livelihoods of millions of farmers. However, conditions have been changing in the delta, with reports of paddy yield stagnation and uncertainty in water availability for cultivation. Understanding the impact of these changes on paddy cultivation is essential to design interventions for adaptation strategies such as crop selection, time of sowing and harvesting, etc. to help farmers adopt climate-resilient practices and eventually contribute towards the goal of SDG-2, Zero Hunger. 


The multiple cropping seasons with small sizes of the croplands in the Delta makes application of readily available LSTR and MSTR satellite data unsuitable for crop type identification in the Delta. Remote sensing of small holder production systems, where cropland sizes are often less than 1 Acre (4046.86 sq m), necessitates the use of HSTR images. However, most HSTR images are expensive and difficult to access, and are usually available with low spectral resolution, which can potentially limit their scope for prediction of the required crop parameters. In this paper, we show that bridging the trade-off between spectral and spatiotemporal resolution in satellite images can improve the crop type classification in a multi-cropping system. We empirically demonstrate that:
\begin{itemize}
    \item High spatiotemporal resolution of satellite images when combined with high spectral resolution statistically improves the performance of the ML models.
    \item High spectral resolution can improve prediction performance in case of low spatial or temporal resolution or both.
\end{itemize}

Existing fusion methods such as pan-sharpening are widely used in remote sensing to merge the high resolution panchromatic band with its corresponding low resolution visible multispectral bands \cite{pan_comparisons}. Recent works propose to fuse a higher number of LR multispectral bands, \cite{sispa_net}, \cite{pan_ICCV}, \cite{pan_multisensor}, but these techniques are limited to availability of panchromatic images.

%% file: sections/data_collection.tex
\section{Data Collection}
We collected data from two sources: field-based surveys and available satellite images. Field data from 172 cropland parcels was collected on crops cultivated in a particular season in the Delta for the years 2018 to 2020. This data includes crop sowing-harvesting period, yield, parcel coordinates and area. Each of the parcel coordinates was then mapped for parcel boundaries using QGIS \cite{QGIS_software}. 

For the satellite data, the Bottom of Atmosphere (BOA) images for Landsat 8 (L8) and Sentinel 2 (S2) were downloaded using GEE Platform \cite{earth_engine} and the Planet Scope (PS)  images from the Planet API \cite{planet}. The BOA images for S2 were only available from 2019, thus it had fewer number of samples. We clipped the image tiles for all the parcels and labelled them based on crop type details using the field data. All the three satellites used here have different sensor characteristics in terms of spatial, temporal and spectral resolutions. L8 has 11 optical spectral bands  with a revisit frequency of 16 days and a spatial resolution of (30-100m). The S2 constellation has a revisit frequency of 5-10 days with 12 bands. Planet Scope is an HSTR satellite having a high spatial resolution of 3m with daily revisits, but with only 4 bands. Apart from these optical characteristics, these satellites also differ in terms of orientation, altitude, equator crossing time, etc. Therefore, we consider the images collected from all these satellites as three separate datasets.

\begin{figure}[hbt!]
     \centering
     \begin{subfigure}[b]{0.4\linewidth}
         \includegraphics[width=\linewidth]{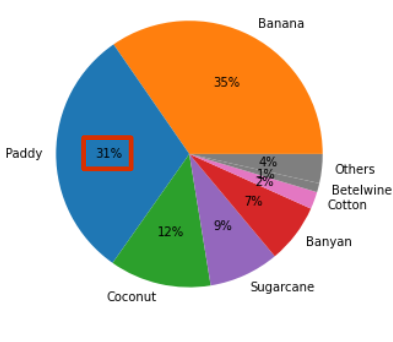}
         \caption{Training Dataset}
         \label{fig:training_distribution}
     \end{subfigure}
     \hfill
     \begin{subfigure}[b]{0.4\linewidth}
         \includegraphics[width=\linewidth]{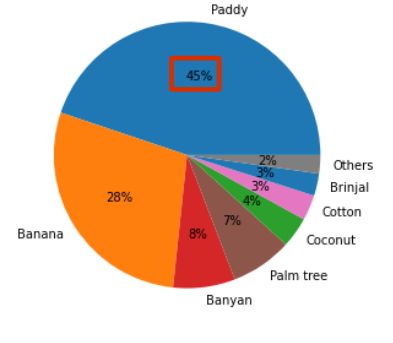}
         \caption{Testing Dataset}
         \label{fig:testing_distribution}
     \end{subfigure}
    \caption{List of different crops and their distribution that are cultivated in the study region, according to the ground survey of 172 parcels. The annual and perennial crops dominate because they have a longer seasonal length. 11 of the 18 different crops contribute only 4\% to the training dataset.}
    \label{fig:data-distribution}
\end{figure}
\begin{figure}[hbt!]
    \centering
    \includegraphics[width=0.7\linewidth]{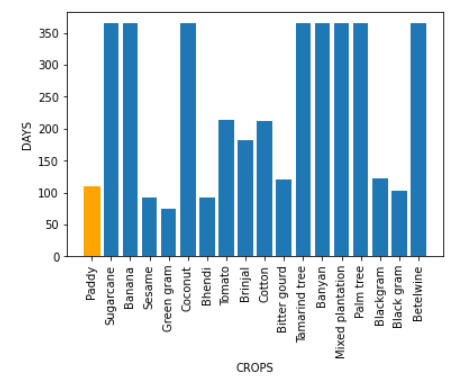}
    \caption{Variations in the length of the growing season for crops cultivated in different cropping seasons. Some crops like sugarcane, banana, etc... are grown throughout the year, while there are many crops that take only a few months to grow.}
    \label{fig:seasonal_length}
\end{figure}

\textbf{Dataset Preparation:} The dataset consists of multiple images for each parcel. Each parcel varies in terms of their spatial extent, geographical location, irrigation source and water availability. Hence, instead of a random train/test split across images, out of the total 172 parcels, we selected 35 spatially and geographically distributed parcels across the study region. \emph{All images} corresponding to these 35 parcels were used to create the test set. The different types of crops and their distribution is shown in Figure \ref{fig:data-distribution}. The high imbalance is due to the difference in growing season length of the crops, as shown in Figure \ref{fig:seasonal_length}. The spatial extent of the parcels is also too small. For MSTR satellites L8 and S2, the average dimensions of the parcels are $3\times3$ and $7\times7$ pixels respectively. Even for the HSTR satellite (PS) the average dimension of the parcels is only $19\times19$ pixels. For crop identification all the images were considered as separate data samples irrespective of the parcel and the cropping season. However, for the time-series based crop classification, we combined all the images corresponding to a parcel for a given cropping season, so that we would have historical data of a parcel for a particular cropping season. The test set in this case was created using the same 35 parcels as before.

%% file: sections/methodology.tex
\section{Methodology}
Due to the relatively small training dataset as well as the significant imbalance in classes, we use a 5-fold cross-validation strategy as opposed to a held-out validation set. In each fold, the validation set is created based on parcels rather than images.  
We use a pretrained Resnet-18 \cite{resnet18} based classifier with two modifications. The input layer contains additional channels to accommodate any additional spectral bands, and the highest residual block is followed by two fully connected layers. To deal with the data imbalance, we make use of the Focal Loss \cite{focal} and the model is trained using the Adadelta optimizer. The results for the best hyperparameter values are reported using the F1-score as the metric.  
The input to the classifier is a batch of multispectral image resized to the average dimensions, i.e. $3\times3$, $7\times7$ and $19\times19$ for L8, S2 and PS respectively.

We adopted the PSE-TAE architecture \cite{psetae} to utilize the spatio-temporal resolution of the satellites. This method, instead of using CNNs to process the input image, uses an unordered set of $n$ pixels sampled randomly from the image. When the number of pixels in the image is less than $n$, an arbitrary pixel from the set is repeated. We chose $n\!=\!9$ for L8, $n\!=\!49$ for S2 and $n\!=\!300$ for PS satellite images. The authors had an equal number of images for all the parcels because the study region and duration were exactly the same for all the parcels. But we are considering the growing season of the crops and hence cannot use the observation dates to compute the positional encoding. Instead, we compute it based on the sequential ordering of the images. The sequence lengths for L8, S2 and PS are 41, 134 and 210 respectively.

%% file: sections/experiments.tex
\section{Experiments}
Tables \ref{tab:spatial}, \ref{tab:spectral}, \ref{tab:spatiotemporal} and \ref{tab:spectral-time-series} show the results for the experiments we conducted to understand the behavior of the model with change in spatial, spatio-spectral, spatio-temporal and spatio-spectro-temporal resolutions respectively. Table \ref{tab:spatial} presents the results when we use the same combination of bands, i.e. red, green and blue, on all the three datasets, to understand the impact of spatial resolution on model performance. Table \ref{tab:spectral} shows the results obtained with different combinations of spectral bands. It summarizes the effects of spectral resolution, by showing the deviation from Table \ref{tab:spatial}, and spatio-spectral resolution by displaying the results of same spectral band combinations with higher spatial resolutions. \ref{tab:spatiotemporal} shows the performance variations using RGB bands for all the datasets, while also considering the temporal dimension. Input dimension represents $T\times H\times W\times B$ tensor, where $T$ denotes the time dimension, $H$ and $W$ denote the height and width of the image and $B$ denotes the number of spectral bands. We use the band combinations that yields best results for each satellite from Table \ref{tab:spectral} to study the effects of spectral resolution with spatio-temporal resolution.

\begin{table}[t!]
    \centering
    \begin{tabular}{c||c|c}
        \hline
        \textbf{Satellite} & \textbf{Spatial Resolution} & \textbf{F1-Score} \\
        \hline
        \hline
        \textbf{Landsat 8} & $3\times3$ & 0.3989 \\
        \textbf{Sentinel 2} & $7\times7$ & 0.5134 \\
        \textbf{Planet Scope} & \textbf{19}$\times$ \textbf{19} & \textbf{0.5184} \\
        \hline
    \end{tabular}
    \caption{Crop identification using RGB bands of the images from different satellites to understand the effects of spatial resolution.}
    \label{tab:spatial}
\end{table}

\begin{table}[t!]
    \centering
    \begin{tabular}{c||c|c}
        \hline
        \textbf{Bands} & \textbf{F1-Score} & \textbf{Gain/Loss}\\
        \hline
        \hline
        \multicolumn{3}{c}{ \textbf{Landsat 8}} \\
        \hline
        G+R+NIR & 0.3549 & \cellcolor{red!20}-0.0440 \\
        R+G+B+SWIR1 & 0.5341 & \cellcolor{green!20}+0.1352\\
        SWIR1+NIR+B & 0.3343 & \cellcolor{red!20}-0.0646\\
        SWIR2+NIR+B & 0.3379 & \cellcolor{red!20}-0.0610\\
        \textbf{NIR+SWIR1+SWIR2} &  \textbf{0.6076} & \cellcolor{green!20} \textbf{+0.2087}\\
        U-B+NIR+SWIR1+SWIR2 & 0.4642 & \cellcolor{green!20}+0.0653\\
        \hline
        \hline
        \multicolumn{3}{c}{ \textbf{Sentinel 2}} \\
        \hline
        G+R+NIR & 0.6027 & \cellcolor{green!20}+0.0893 \\
        R+G+B+SWIR1 & 0.5904 & \cellcolor{green!20}+0.0770\\
        SWIR1+NIR+B & 0.6089 & \cellcolor{green!20}+0.0955\\
        SWIR2+NIR+B & 0.6113 & \cellcolor{green!20}+0.0979\\
        \textbf{NIR+SWIR1+SWIR2} &  \textbf{0.6122} & \cellcolor{green!20} \textbf{+0.0988} \\
        R+G+B+RED-EDGE2 & 0.593 & \cellcolor{green!20}+0.0796\\
        \hline
        \hline
        \multicolumn{3}{c}{ \textbf{Planet Scope}} \\
        \hline
        G+R+NIR & 0.3892 & \cellcolor{red!20}-0.1292 \\
        \textbf{R+G+B+NIR} &  \textbf{0.5455} & \cellcolor{green!20} \textbf{+0.0272} \\
        \hline
    \end{tabular}
    \caption{Crop identification using different spectral band combinations of the images from different satellites to understand the effects of spatio-spectral resolution. Gain/Loss represent the deviation of the results from the results obtained using RGB band combination for the same satellite.}
    \label{tab:spectral}
\end{table}

\begin{table}[t!]
    \centering
    \begin{tabular}{c||c|c}
        \hline
        \textbf{Satellite} & \textbf{Input Dimension} & \textbf{F1-Score} \\
        \hline
        \hline
        \textbf{Landsat 8} & $41\times3\times3\times3$ & 0.9167 \\
        \textbf{Sentinel 2 *} & $134\times7\times7\times3$ & 0.7692 \\
        \textbf{Planet Scope} & \textbf{210}$\times$ \textbf{19}$\times$ \textbf{19}$\times$ \textbf{3} & \textbf{0.9268} \\
        \hline
    \end{tabular}
    \caption{Time-series crop classification using RGB bands of the images from different satellites to understand the effects of spato-temporal resolution. * Sentinel 2 had images for only 2019-2020, therefore when the images were combined to create a time-series dataset the number of samples were reduced significantly.}
    \label{tab:spatiotemporal}
\end{table}

\begin{table}[t!]
    \centering
    \begin{tabular}{c||c|c}
        \hline
        \textbf{Satellite} & \textbf{Input Dimension} & \textbf{F1-Score} \\
        \hline
        \hline
        \textbf{Landsat 8} & \textbf{41}$\times$ \textbf{3}$\times$ \textbf{3}$\times$ \textbf{4} & \textbf{0.9388} \\
        \hline
        \textbf{Planet Scope} & $210\times19\times19\times4$ & 0.9271 \\
        \hline
    \end{tabular}
    \caption{Effect of spectral resolution on time-series classification}
    \label{tab:spectral-time-series}
\end{table}

%% file: sections/results.tex
\section{Results and Analysis}
When we consider only the spatial context of the images we get the best results using the HR images of Planet Scope, even though the results are not acceptable for field deployment. This poor performance can possibly be explained due to different labels being assigned to two visually similar image sets. Images from parcels of paddy class and non-paddy class would have a very similar visual appearance closer to the sowing dates when the crops' phenological signature is largely absent. When PS images from only the second half of the cropping season were used to train the model, an F1-score of 0.5964 was observed on the test set. 
With a time-series crop classification model, we observed a similar pattern that better prediction performance is obtained as we move from MSTR to HSTR images. As the train set for S2 was too small, we were unable to train models well. 

Tables \ref{tab:spectral} and \ref{tab:spectral-time-series} show the results for different combinations of spectral bands. We report the F1-score on the test dataset along with the corresponding gain/loss as compared to results obtained using RGB band combination of that satellite. We observed that naively combining all the spectral bands degraded the prediction performance, however, upon using different combinations of bands as suggested in \cite{band1}, \cite{band2} and \cite{band3}, the multispectral information could substantially improve the performance of the ML models. 
For certain band combinations, we observed that high spectral resolution along with MSTR images can even outperform the results obtained using RGB bands of HSTR images. Using the bands that gave best results, we further validated the hypotheses with a time-series classification problem. We again observe that multispectral information outperforms its own baseline results. Since PS has only four bands, i.e. Red, Green, Blue and NIR, it therefore cannot use the spectral bands that gave best results on Landsat and Sentinel. Based on these observations, we claim that the high spectral resolution of MSTR satellite images improved the performance of the ML models over those that used the HSTR images. We hypothesize that the performance of PS could have been even better if it had higher spectral resolution.

%% file: sections/conclusion.tex
\section{Conclusions and Future Work}
In this paper we demonstrated how higher spectral resolution improves the effectiveness of the model even when we don't have HSTR images. Many space agencies are making advancements towards improving the sensor characteristics, which also includes narrowing the bandwidth and increasing the numbers of the bands. But many domains, such as agronomy, rely on historical data to study the pattern throughout the years. We definitely need advancements in the sensors, but we also want historical data to be improved and to be consistent with the current advancements. 

\cite{rgb2nir} and \cite{synthetic_nir} propose to generate new spectral bands, but they need ground truth band values. Recent work on pan-sharpening is interesting \cite{sispa_net} and \cite{pan}, however, here the approach is to increase spatial resolution of LR multispectral images using the HR panchromatic images. An interesting research direction could be to investigate cross-satellite fusion to achieve high spectral resolution for HSTR images. 